\documentclass[abstract]{nldl}

\renewcommand{\DOI}{12345}

\usepackage[utf8]{inputenc}
\usepackage{url}
\usepackage{graphicx}
\usepackage{authblk}

\usepackage{algorithm}
\usepackage{algorithmic}

\usepackage[square,numbers]{natbib}

\usepackage{microtype}
\usepackage{tikz}
\usepackage{pgf}
\usetikzlibrary{shapes, arrows}

\usepackage[hidelinks]{hyperref}

\title{Leveraging Topological Maps in Deep Reinforcement Learning for Multi-Object Navigation}

\author{Simon Hakenes\thanks{Corresponding Author: \href{mailto:simon.hakenes@ini.rub.de}{simon.hakenes@ini.rub.de}} }
\author{Tobias Glasmachers}
\affil{Institute for Neural Computation, Ruhr University Bochum, Germany}

\date{\vspace{-5ex}}

\begin{document}
\nldlmaketitle

\begin{abstract}
This work addresses the challenge of navigating expansive spaces with sparse rewards through Reinforcement Learning (RL).
Using topological maps, we elevate elementary actions to object-oriented macro actions, enabling a simple Deep Q-Network (DQN) agent to solve otherwise practically impossible environments. 
\end{abstract}

\section{Introduction}

Navigating large spaces is practically impossible for standard RL algorithms relying on elementary actions.
While random exploration is theoretically possible, it proves practically infeasible in large state spaces.
Solutions incorporating scene understanding and prior knowledge about space to design macro actions are a promising direction. 
Instead of leaving it up to the RL algorithm to learn these macro actions solely based on the information from the Markov Decision Process \cite{Sutton1999-yp}, incorporating external world knowledge about space is a more efficient strategy for designing macro actions.

We propose using objects to design macro actions. 
Translation between macro and elementary actions is done with a topological map where object positions and their connections are stored. 
In contrast to metric maps, topological maps are a more effective planning tool and are even employed by mammals for spatial navigation \cite{Parra-Barrero2023-xm}.

All the ingredients for such a system are already developed: 
For instance, Simultaneous Localization and Mapping (SLAM) \cite{Mur-Artal2015-cb} can build a map of 3D space from monocular RGB(D) data while localizing the agent on the map. 
Storing traversed paths as a graph is straightforward, and with it building a topological map. 
Finding the shortest path in a graph is solved for a long time. 
Object detection and recognition algorithms grant us semantic understanding of our visual environment and RL algorithms leverage a single sparse and delayed reward signal to select actions. 
The challenge lies in cohesively integrating these components.

\begin{figure}
    \centering
    \includegraphics[width=0.32\linewidth]{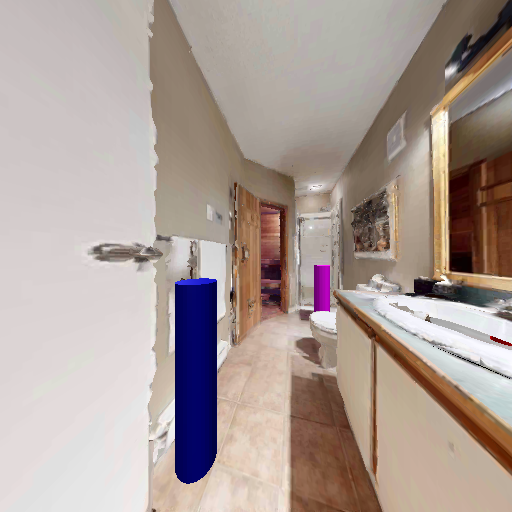}
    \includegraphics[width=0.32\linewidth]{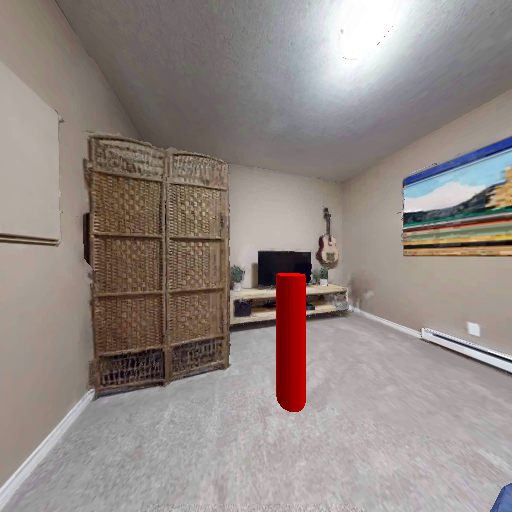}
    \includegraphics[width=0.32\linewidth]{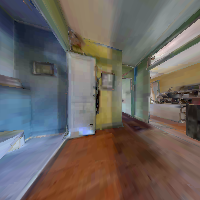}
    \caption{Screenshots of the environment. The cylinders mark the goals.}
    \label{fig:env_screenshots}
\end{figure}

In this work, we integrate the above mentioned components and connect it to a novel network architecture and training scheme for the DQN \cite{Mnih2015-ut} inspired by \cite{He2016-gy}. 
An object oriented topological map is created and connected to the DQN to enable efficient exploration and navigation.
Recognizing the inefficiencies of RL, we aim to streamline as much of the process as possible, minimizing the workload for the core RL algorithm.

Existing literature on navigation can be broadly categorized into differentiable metric memory \cite{Parisotto2018-an,Zhang2017-ay}, non-differential metric maps \cite{Wani2020-km} and unstructured and implicit memory (map less memory) \cite{Mnih2016-hl}.
Some authors use topological maps \cite{Savinov2018-pg}, although their task is simpler and not trained by RL. 

To the best of our knowledge, this is the first work about navigation where the agent remains oblivious about what the goals are (e.g., target images or distance-based rewards), and their sequential order. 
We leverage biologically plausible topological maps and and rely exclusively on reinforcement learning for training.

\section{Method}
Whenever an object is detected in an RGBD frame, its position is estimated from the agent's position, viewing angle and its depth. 
A node on the map is created, which stores the corresponding pixels of the object inside the bounding box, as well as the position and a flag whether or not the node is already explored, greatly aiding exploration.
Nodes are marked as explored when the agent was close to the node. 
An edge is added if the agent moved from one node to the other, creating a navigable graph.

As the map grows, so does the action space.
Necessarily, we changed the DQN architecture and training procedure, inspired by \cite{He2016-gy}:
The neural network, implemented as a convolutional neural network, takes the outer product of one action feature (in the form of pixels) and a one-hot encoded state vector indicating task progress as input. 
In each step, Q-Values---representing expected future rewards---are computed iteratively for all map-stored objects.
The agent then selects the action with the highest Q-Value. 
Unexplored actions receive a Q-Value bonus to encourage exploration.

\section{Experiments and Results}
We use the Habitat 2.0 environment \cite{Savva2019-hh, Szot2021-ti} combined with the photo realistic Habitat-Matterport 3D dataset \cite{Ramakrishnan2021-cz} (Fig.~\ref{fig:env_screenshots}). 
Based on \cite{Wani2020-km}, the task requires to find up to 3 target objects in a scene sequentially. 
While each episode retains the object order, their positions vary. 
The target objects are color coded cylinders. 
In a more challenging variant the objects are real world objects that blend into the scene and are not recognizable just by their color.

Each scene includes multiple rooms and hundreds of different objects. 
The only cue is the reward signal, which makes it a challenging environment. 
Two reward systems were tested: one provided a $+1$ reward for each identified subgoal, while the more challenging variant granted $+1$ only when all goals were achieved.

The experiments are conducted in 100 different scenes with different goals positions for each scene. 
For now, a working SLAM and object detection are assumed. In the Habitat environment, they are not needed since corresponding ground truth information is available.

\begin{figure}
    \centering
    \input{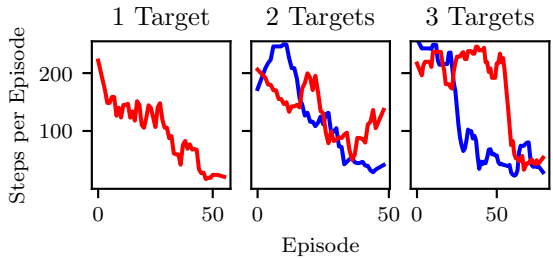}
    \caption{Plots showing steps per episode for 1, 2, and 3 targets. Blue lines depict intermediate reward experiments, and red lines represent runs with only a single reward at the episode's end. There is a limit of 250 macro actions per episode.}
    \label{fig:results}
\end{figure}

As shown in Fig.~\ref{fig:results} there is a decrease in episode steps length {during training, which shows that the agent is capable to (re-)recognize the objects and to learn the correct order reliably.

\section{Conclusion and Future Work}
The results clearly show the great potential of using macro actions based on topological maps in RL. 
Current limitations include long training, reliance on handcrafted map heuristics and the need for effective object detection and SLAM algorithms. 

For future work, we aim to use photo-realistic targets that blend into the scene, incorporate an autoencoder to assist the training of the convolutional layer as in \cite{Hakenes2019-tg}, integrate an actual SLAM algorithm, introduce more targets and investigate different network architectures.

\bibliographystyle{abbrvnat}
\bibliography{references}

\end{document}